# BRAIN TUMOR MRI IMAGE CLASSIFICATION WITH FEATURE SELECTION AND EXTRACTION USING LINEAR DISCRIMINANT ANALYSIS

V.P.Gladis Pushpa Rathi[1] and Dr.S.Palani[2]

[1]Department of Computer Science and Engineering, Sudharsan Engineering College Sathiyamangalam, Pudukkottai , India

gladispushparathi@gmail.com

[2]Department of Electronics and Communication Engineering , Sudharsan Engineering College Sathiyamangalam, Pudukkottai , India
palani_keeranur@yahoo.co.in

## ABSTRACT

*Feature extraction is a method of capturing visual content of an image. The feature extraction is the process to represent raw image in its reduced form to facilitate decision making such as pattern classification. We have tried to address the problem of classification MRI brain images by creating a robust and more accurate classifier which can act as an expert assistant to medical practitioners. The objective of this paper is to present a novel method of feature selection and extraction. This approach combines the Intensity, Texture, shape based features and classifies the tumor as white matter, Gray matter, CSF, abnormal and normal area. The experiment is performed on 140 tumor contained brain MR images from the Internet Brain Segmentation Repository. The proposed technique has been carried out over a larger database as compare to any previous work and is more robust and effective. PCA and Linear Discriminant Analysis (LDA) were applied on the training sets. The Support Vector Machine (SVM) classifier served as a comparison of nonlinear techniques Vs linear ones. PCA and LDA methods are used to reduce the number of features used. The feature selection using the proposed technique is more beneficial as it analyses the data according to grouping class variable and gives reduced feature set with high classification accuracy.*

## 1. INTRODUCTION

Brain tumors are abnormal and uncontrolled proliferations of cells. Some originate in the brain itself, in which case they are termed primary. Others spread to this location from somewhere else in the body through metastasis, and are termed secondary. Primary brain tumors do not spread to other body sites, and can be malignant or benign. Secondary brain tumors are always malignant. Both types are potentially disabling and life threatening. Because the space inside the skull is limited, their growth increases intracranial pressure, and may cause edema, reduced blood flow, and displacement, with consequent degeneration, of healthy tissue that controls vital functions. Brain tumors are, in fact, the second leading cause of cancer-related deaths in children and young adults. According to the Central Brain Tumor Registry of the United States (CBTRUS), there will be 64,530 new cases of primary brain and central nervous system tumors diagnosed by the end of 2011. Overall more than 600,000 people currently live with the disease. [2]

Early and accurate diagnosis of brain tumor is the key for implementing successful therapy and treatment planning. However the Diagnosis is a very challenging task due to the large variance and complexity of tumor characterization in images, such as size, shape, location and intensities and can only be performed by professional neuro radiologists. In the recent past several research works have been done for the diagnosis and treatment of brain tumor. The most important advantage of MR imaging is that it is non-invasive technique.

The use of computer technology in medical decision support is now widespread and pervasive across a wide range of medical area such as cancer research, gastroenterology, brain tumors etc. MRI is the viable option now for the study of tumor in soft tissues. The method clearly finds tumor types, size and location. MRI is a magnetic field which builds up a picture and has no known side effects related to radiation exposure. It has much higher details in soft tissues. Researcher had proposed various features for classifying tumor in MRI. The statistical, Intensity, Symmetry, Texture features etc, which utilize gray value of tumors are used here for classifying the tumor. However the gray values of MRI tend to change due to over – enhancement or in the presence of noise.[4]

In image processing, feature extraction is a special form of dimensionality reduction. When the input data to an algorithm is too large to be processed and it is suspected to be notoriously redundant (much data, but not much information) then the input data will be transformed into a reduced representation set of features (also named features vector). Transforming the input data into the set of features is called feature extraction. If the features extracted are carefully chosen it is expected that the features set will extract the relevant information from the input data in order to perform the desired task using this reduced representation instead of the full size input.[3]

This paper presents a novel approach for feature extraction and selection. Feature extraction involves simplifying the amount of resources required to describe a large set of data accurately. When performing analysis of complex data, one of the major problems stems from the number of variables is involved. Analysis with a large number of variables generally requires a large amount of memory and computation power or a classification algorithm which over fits the training sample and generalizes poorly to new samples. Feature extraction is a general term for methods of constructing combinations of the variables to get around these problems while still describing the data with sufficient accuracy.

Feature selection is the technique of selecting a subset of relevant features for building robust learning models by removing most irrelevant and redundant features from the data, feature selection helps improve the performance of learning models by:

- Alleviating the effect of the curse of dimensionality.
- Enhancing generalization capability.
- Speeding up learning process.
- Improving model interpretability.

Feature selection also helps people acquire better understanding about their data by telling them which are the important features and how they are related with each other. In the proposed method by using PCA+ LDA, we obtain a combining process for feature reduction. The first processing step is PCA transformation without dimension reduction, in other words, all the eigenvalues are kept in a matrix. Then numbers of eigen values, which have highest and effective values, are computed. The average cumulative sum of the eigenvalues, obtained from PCA, is depicted against the number of eigenvalues. It shows that the sum of two largest eigenvalues has the value of 99.99 percentages of the whole eigenvalues. This means that the third eigenvalue will not affect the results. Therefore, we have an action of LDA in second step where feature matrix dimensionality reduction discounts features from 15 to 2. Limiting the feature vectors by such a combining process leads to an increase in accuracy rates and a decrease in complexity and computational time.

This Paper is organized as follows. Section 2 describes the related works .In section 3 we describe normalization, and feature extraction , selection and comparative analysis of PCA and LDA In section 4 tumor classification and experimental results are discussed. The conclusions are given in section 5.

## 2. RELATED WORKS

For the diagnostic process in pathology, we can discern two main steps. First pathologists observe tissue and recognize certain histological attributes related to the degree of tumor malignancy. In a second step interpret their histological findings and come up with a decision related to tumor grade. In most of the cases, pathologists are unaware of precisely how many attributes have been considered in their decision but they are able to classify tumors almost instantly and unconscious of the complexity of the task performed.

Pathologists are capable to verbalize their impression of particular features. For example, they can call mitosis and apoptosis as "present" or "absent" but they do not know how precisely these concepts have to be taken into account in the decision process. To this end, although the same set of features is recognized by different histopathologists, each one is likely to reach to a different diagnostic output. To confine subjectivity, considerable efforts have been made based on computer-assisted methods with a considerable high level of accuracy. It proposes data-driven grading models such as statistical vector machines, artificial neural networks, and decision trees coupled with image analysis techniques to incorporate quantitative histological features.

However, besides the retention and enhancement of achieved diagnostic accuracies in supporting medical decision, one of the main objectives, is to enlarge the inter-operability and increase transparency in decision-making. The latter is major importance in clinical practice, where a premium is placed on the reasoning and comprehensibility of consulting systems.

A number of approaches have been used to segment and predict the grade and volume of the brain tumor. EI papageevgious et.al (applied soft computing 2008) in their work proposed a fuzzy cognitive map (FCM) to find the grade value of tumor. Authors used the soft computing method of fuzzy cognitive maps to represent and model expert's knowledge FCM grading model achieved a diagnostic output accuracy of 90.26% & 93.22 % of brain tumors of low grade and high grade respectively. They proposed the technique only for Characterization and accurate determination of grade [1].

Shafab Ibrahim, Noor Elaiza in their work proposed an implementation of evaluation method known as image mosaicing in evaluating the MRI brain abnormalities segmentation study. 57 mosaic images are formed by cutting various shapes and size of abnormalities and pasting it onto normal brain tissue. PSO, ANFIS, FCM are used to segment the mosaic images formed. Statistical analysis method of receiver operating characteristic (ROC) was used to calculate the accuracy [7].

S.Karpagam, S.Gowri, in their work proposed detection of tumor growth by advanced diameter technique using MRI data. To find the volume of brain tumor they proposed diameter and graph based methods. The result shows tumor growth and volume [8].

Matthew C.clrk Lawrence et.al proposed a system that automatically segments and lables tumor in MRI of the human brain. They proposed a system which integrates knowledge based techniques with multispectral analysis. The results of the system generally correspond well to ground truth, both on a per state basis and more importantly in tracking total volume during treatment over time [5].

Carlos A.Patta, Khan IbleKharuddin and Robert, in their work suggested a enhanced implementation of artificial neural network algorithm to perform segmentation of brain MRI data learning vector quantization and is used for segmentation. Their result suggests excellent brain tissue segmentation [6].

In this paper a new and improved method is implemented by combining LDA & PCA for feature reduction and SVM is used for classification of MRI images. Compared to the previous work suggested in the literature discussed above high accuracy is achieved for feature selection and extraction.

## 3. PROPOSED METHOD

The architecture of our system is illustrated in Figure 1. The major components of our system are Brain tumor Database, Normalisation, Feature selection, Feature extraction and Classification.

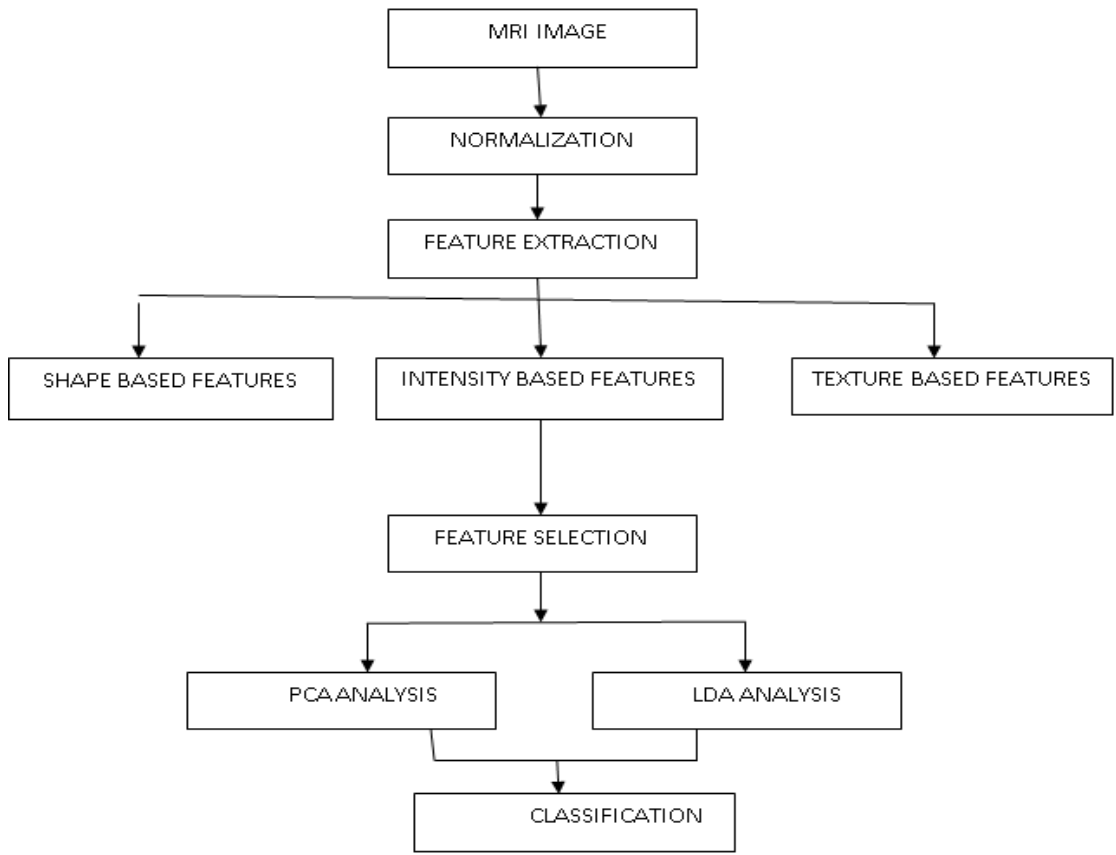

Figure 1. Architecture of proposed method

### 3.1. Data Description

Experiments are conducted on MR images collected from 20 different patients with gliomas. Each patient has 3 sequences of MR images T1, T2 and FLAIR. Each volume contains 24 slices in axial plain with 5 mm slice thickness. MR imaging was performed on 3.0T Siemens devices. The imaging conditions of different protocols are; T1 weighted, T2 weighted, and Flair weighted. The MRI image data description of the proposed method is shown in table 1. Each set of features are individually normalized to the range of 0 to 255.

Table 1. Data Description

| Attribute | Description | Value |
|---|---|---|
| Age | Age in Years | 17 to 83 |
| Sex | Sex | Men -46, Women -52 |
| Matrix size | Size of the matrix | 192*256*192 |
| Voxel size | Size of the voxel | 0.98*0.98*1mm |
| Sequences | MRI image sequences | Axial 3D T1 weighted , Sagittal 3D T2 weighted , Fluid Attenuated Inversion Recovery (FLAIR) |

### 3.2. Normalization

Initially, these MRI images are normalized to gray level values from 0 to 1 and the features are extracted from the normalized images. Since normalization reduces the dynamic range of the intensity values, feature extraction is made much simpler.

### 3.3. Feature Extraction

Features, the characteristics of the objects of interest, if selected carefully are representative of the maximum relevant information that the image has to offer for a complete characterization of a lesion. Feature extraction methodologies analyse objects and images to extract the most prominent features that are representative of the various classes of objects. Features are used as inputs to classifiers that assign them to the class that they represent. The purpose of feature extraction is to reduce the original data by measuring certain properties, or features, that distinguish one input pattern from another pattern. The extracted feature should provide the characteristics of the input type to the classifier by considering the description of the relevant properties of the image into feature vectors. In this proposed method we extract the following features.

Shape Features   - circularity, irregularity, Area, Perimeter, Shape Index

Intensity features – Mean, Variance, Standard Variance, Median Intensity, Skewness, and Kurtosis

Texture features  –Contrast, Correlation, Entropy, Energy, Homogeneity, cluster shade, sum of square variance.

Accordingly, 3 kinds of features are extracted, which describe the structure information of intensity, shape, and texture. These features certainly have some redundancy, but the purpose of this step is to find the potential by useful features. In the next step the feature selection will be performed to reduce the redundancy.

## 3.4. Feature Selection

Feature selection (also known as subset selection) is a process commonly used in machine learning, wherein a subset of the features available from the data is selected for application of a learning algorithm. The best subset contains the least number of dimensions that contributes to high accuracy; we discard the remaining, unimportant dimensions.

### 3.4.1. Forward Selection

This selection process starts with no variables and adds them one by one, at each step adding the one that decreases the error the most, until any further addition does not significantly decrease the error. We use a simple ranking based feature selection criterion, a two –tailed t-test, which measures the significance of a difference of means between two distributions, and therefore evaluates the discriminative power of each individual feature in separating two classes.

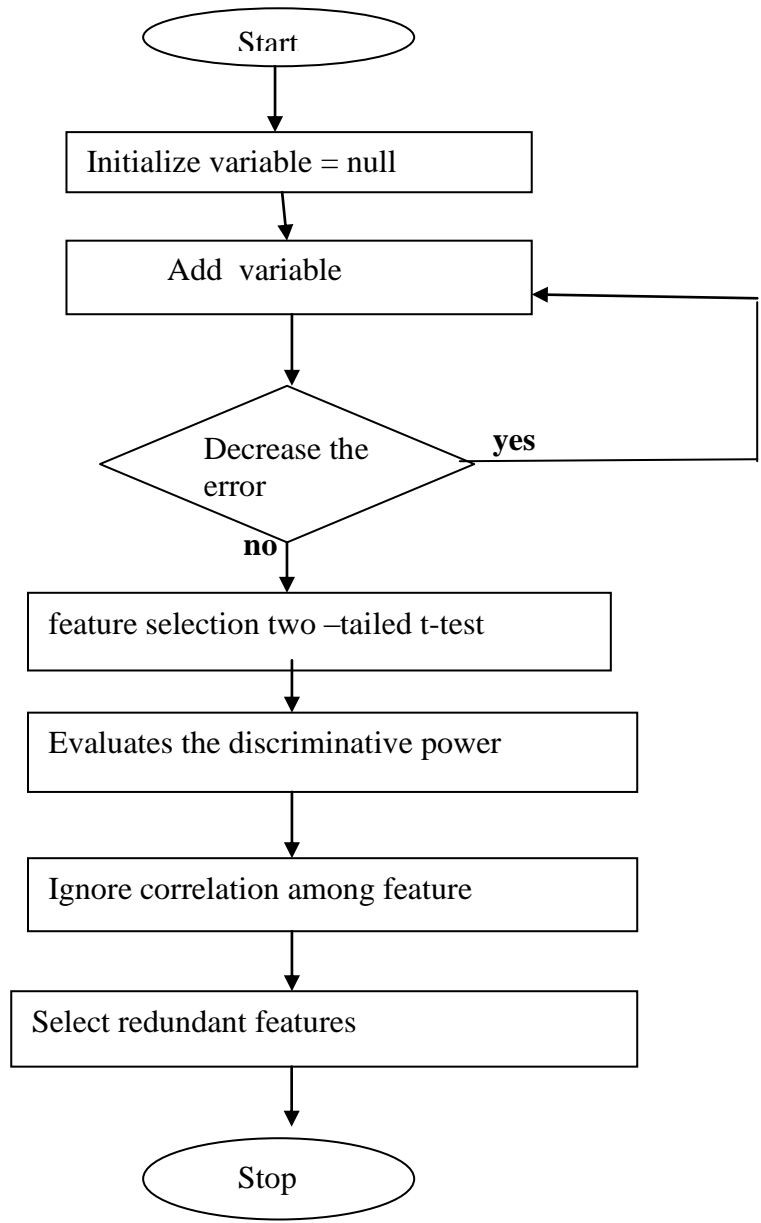

Figure 2. Steps for forward selection

The features are assumed to come from normal distributions with unknown, but equal variances. Since the correlation among features has been completely ignored in this feature ranking method, redundant features can be inevitably selected, which ultimately affects the classification results. Therefore, we use this feature ranking method to select the more discriminative feature, e.g.by applying a cut-off ratio (p value<0.1), and then apply a feature subset selection method on the reduced feature space, as detailed below. Figure 2 shows the procedure for forward selection

### 3.4.2. Backward Selection

This selection process starts with all the variables and removes them one by one, at each step removing the one that decreases the error the most (or increases it only slightly), until any further removal increases the error significantly. To reduce over fitting, the error referred to above is the error on a validation set that is distinct from the training set. The support vector machine recursive feature elimination algorithm is applied to find a subset of features that optimizes the performance of the classifier. This algorithm determines the ranking of the features based on a backward sequential selection method that remove one feature at a time. At each time, the removed feature makes the variation of SVM based leave-one-out error bound smallest, compared to removing other features

### 3.5. Classification

There are many possible techniques for classification of data. Principal Component Analysis (PCA) and Linear Discriminant Analysis (LDA) are the two commonly used techniques for data classification and dimensionality reduction. Linear Discriminant Analysis easily handles the case where the within-class frequencies are unequal and their performance has been examined on randomly generated test data. This method maximizes the ratio of between-class variance to the within-class variance in any particular data set thereby guaranteeing maximal separability. The use of Linear Discriminant Analysis for data classification is applied to classification problem in speech recognition We decided to implement an algorithm for LDA in hopes of providing better classification compared to Principal Components Analysis. The prime difference between LDA and PCA is that PCA does more of feature classification and LDA does data classification. In PCA, the shape and location of the original data sets change when transformed to a different space whereas LDA doesn't change the location but only tries to provide more class separability and draw a decision region between the given classes. The classification process is divided into the training phase and the testing phase. In the training phase known data are given. In the testing phase, unknown data are given and the classification is performed using the classifier after training. The accuracy of the classification depends on the efficiency of the training.

### 3.5.1. Principal Component Analysis

Principal components are the projection of the original features onto the eigenvectors and correspond to the largest eigenvalues of the covariance matrix of the original feature set. Principle components provide linear representation of the original data using the least number of components with the mean squared error minimized

PCA can be used to approximate the original data with lower dimensional feature vectors. The basic approach is to compute the eigenvectors of the covariance matrix of the original data, and approximate it by a linear combination of the leading eigenvectors. By using PCA procedure, the test image can be identified by first, projecting the image onto the eigen space to obtain the corresponding set of weights, and then comparing with the set of weights of the faces in the training set.

The problem of low-dimensional feature representation can be stated as follows:

Let $X=(x_1, x_2, x_3, x_4 \ldots x_i \ldots x_n)$ represent the n×N data matrix, where each xi is a face vector of dimension n, concatenated from a p×q face image. Here n represents the total number of pixels(p,q) in the face image and N is the number of face images in the training set. The PCA can be considered as a linear transformation from the original image vector to a projection feature vector, i.e

$$Y = W^T X \qquad (1)$$

where Y is the m×N feature vector matrix, m is the dimension of the feature vector, and transformation matrix W is an n×m transformation matrix whose columns are the eigenvectors corresponding to the m largest eigen values computed using equation(2)

$$\lambda e_i = S e_i \qquad (2)$$

where $e_i$ and $\lambda$ are eigenvectors and eigen values of the matrix respectively. Here the total scatter matrix S and the mean image of all samples are defined as

$$s = \sum_{i=1}^{N} (x_i - \mu)(x_i - \mu)^T, \quad \mu = 1/N \sum_{i=1}^{N} x_i \qquad (3)$$

after applying the linear transformation $W^T$ the scatter of the transformed feature vectors { y1,y2,…..yN} is $W^TSW$. In PCA, the projection $W_{opt}$ is chosen to maximize the determinant of the total scatter matrix of the projected samples, i.e.,

$$W_{opt} = \arg^{MAX-w} |W^T S W| = [w_1, w_2, \ldots w_m] \qquad (4)$$

where { $w_i, i=1,2,\ldots m$} is the set of n-dimensional eignvectors of S corresponding to the m largest eigen values. In other words, the input vector (face) in an n-dimensional space is reduced to a feature vector in an m- dimensional subspace. We can see that the dimension of the reduced feature vector m is much less than the dimension of the input faces vector n.

### 3.5.2. Linear Discriminant Analysis

LDA methods are used in statistics, pattern recognition, and machine learning to find a linear combination of features. LDA attempts to express 1ess one dependent variable as a linear combination of other features or measurements. LDA is also closely related to PCA and factor analysis in that they both look for linear combination of variables which best explain the data. LDA explicitly attempts to model the difference between the classes of data. PCA on the other hand does not take into account of any difference in class, and factor analysis builds the feature. Combination is based on differences rather than similarities. LDA searches for those vectors in the underlying space that best discriminable among classes. More formally given a number of independent features relative to which the data is described, LDA creates a linear combination of those which yields the largest mean differences between the desired classes. We define two measures: 1) one is called within- class scatter matrix as given by

$$S_w = \sum_{j=1}^{c} \sum_{i=1}^{N_j} \left(x_i^j - \mu_j\right)\left(x_i^j - \mu_j\right)^T \qquad (5)$$

where $x_i^j$ is the $i^{th}$ sample of class j, $\mu_j$ is the mean of class j, c is the number of classes, and $\mu_j$ is the number of samples in class j and 2)between class scatter matrix

$$Sb = \sum_{j=1}^{c} (\mu_j - \mu)(\mu_j - \mu)^T \qquad (6)$$

where µ represents the mean of all classes.

### 2.5.2. Support Vector Machine

Support vector machines are a state of the art pattern recognition technique grown up from statistical learning theory. The basic idea of applying SVMs for solving classification problems can be stated briefly as follows: a) Transform the input space to higher dimension feature space through a non-linear mapping function and b) Construct the separating hyperplane with maximum distance from the closest points of the training set.

In the case of linear separable data, the SVM tries to find among all hyper planes that minimize the training error, the one that separates the training data with maximum distance from their closest points

$$w \bullet x + b = 0 \qquad (7)$$

with w and b are weight and bias parameters respectively.

In order to define the maximal margin hyperplane (MMH) the following constrains must be fulfilled:

$$\text{Minimize } \tfrac{1}{2} \|w\|^2 \text{ with } y_i(w \bullet x_i + b) \geq 1 \qquad (8)$$

This is a classic nonlinear optimization problem with inequality constraints. It can be solved by the karush-kuhn-Tucker (KKT) theorem by introducing Lagrange multipliers

$$\text{maximize } \sum_{i=1}^{l} a_i - \frac{1}{2} \sum_{i,j=1}^{l} y_i y_j a_i a_j x_i^T x_j \qquad (9)$$

$$\text{subject to } \sum_{i=1}^{l} a_i y_i = 0 \text{ and } a_i \geq 0 \qquad (10)$$

The solution of w is:

$$w = \sum_{i=1}^{l} a_i y_i x_i \qquad (11)$$

The only nonzero solutions define those training data (usually a small percentage of the initial data set) that are necessary to form the MMH and are called support vectors. The optimal hyper plane theory is generalized for non-linear overlapping data by the transformation of the input vectors into a higher dimensional feature space through a mapping function

$$x_i \in R^n \rightarrow z(x) = [a_1 \Phi_1(x), a_2 \Phi_2(x), \ldots a_n \Phi_n(x)]^T \in R^f \qquad (12)$$

The KKT conditions transform to

$$\text{Maximize } \sum_{i=1}^{l} a_i - \frac{1}{2} \sum_{i,j=1}^{l} y_i y_j a_i a_j K(x_i x_j) \qquad (13)$$

$$\text{Subject to} \sum_{i=1}^{l} a_i y_i = 0 \text{ and } a_i \geq 0 \qquad (14)$$

The optimization problem is solved using the MATLAB optimization toolbox

## 4. EXPERIMENT RESULTS

In all the selected 60 features, there are 22 Intensity based features, 5 Shape based features, 33 texture based features. It is found that there are 3 kinds of features extracted in our work and are all useful for the classification. Besides, the distribution of T1, T2, and FLAIR are 10, 20,30 respectively. It means FLAIR provides the most information for tumor segmentation, T2 provides less and T1provides the least. This result is in accordance with the conclusion in Medical Imaging that FLAIR and T2 are more sensitive in pathological discrimination than T1. The distribution of selected features is shown in table 2.

Table 2: Distribution of Selected Features

| Features | T1 | T2 | FLAIR | TOTAL |
|---|---|---|---|---|
| Intensity | 6 | 5 | 11 | 22 |
| Shape | 1 | 1 | 3 | 5 |
| Texture | 8 | 5 | 20 | 33 |
| Total | 10 | 20 | 30 | 60 |

Efficiency or accuracy of the classifiers for each texture analysis method is analysed based on the error rate. This error rate can be described by the terms true and false positive and true and false negative as follows:

**True Positive (TP):** The test result is positive in the presence of the clinical abnormality.
**True Negative (TN):** The test result is negative in the absence of the clinical abnormality.
**False Positive (FP):** The test result is positive in the absence of the clinical abnormality.
**False Negative (FN):** The test result is negative in the presence of the clinical abnormality

$$FP = \text{false positive pixels number /tumor size} \qquad (15)$$

$$FN = \text{false negative pixel number / tumor size} \qquad (16)$$

$$\text{Correct rate} = FP + FN \qquad (17)$$

Figure 3 shows the result of pre-processed image details original, blurred, edge detection and segmented images . The average correct rate by the method presented is 97.82% with FP of 1.0% and FN of 2.50%. All the features produce classification accuracy of 98.87% using LDA. The extracted four PCA components are classified using LDA and SVM classification and the accuracy achieved is 96%. . The overall accuracy percentage details are shown in fig 4. The comparative analysis of the proposed method and the existing algorithms are shown in table 3. Comparative analysis of the proposed method and the existing systems are shown in figure 5.

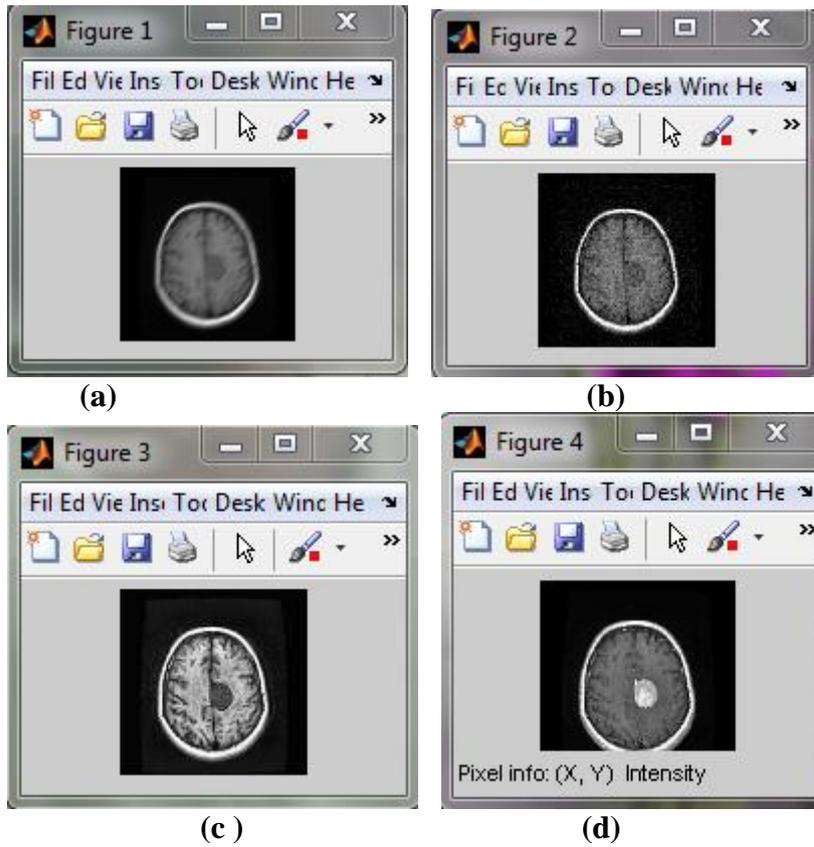
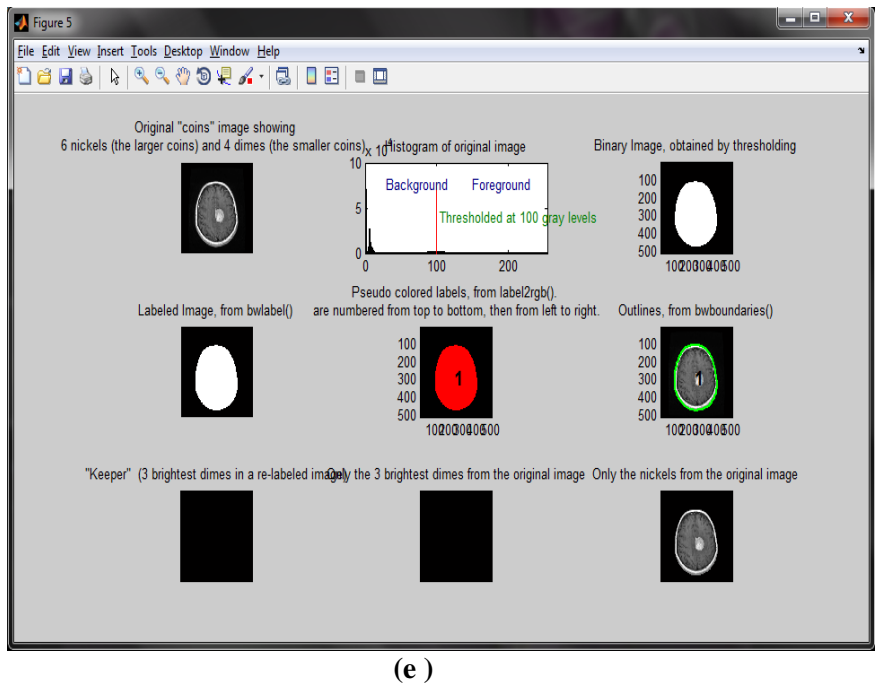

Fig: 3 Pre-processing results a) Original image b) blurred image c) edge detection d) Segmentation e) normalization and all process

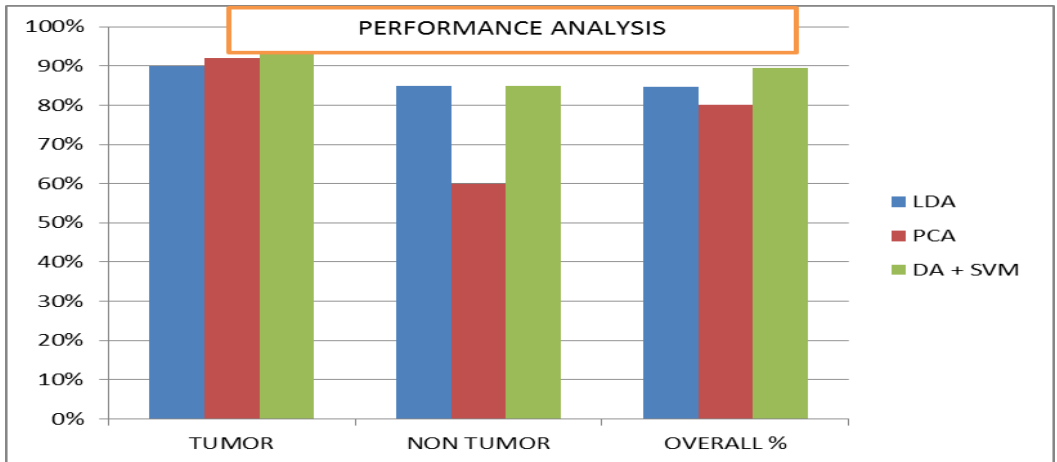

Fig: 4 overall accuracy performance of the proposed method

Table 3.Comparative analysis

| Classification accuracy | FP | FN | Correct rate | With FS | Without FS |
|---|---|---|---|---|---|
| Proposed method | 1.00% | 2.50% | 97.82% | 98.87% | 98.77% |
| KNN | 2.75% | 7.51% | 93.50% | 98.48% | 95.47% |
| Fuzzy connectedness | 2.95% | 5.02% | 92.04% | 98.35% | 97.47% |
| AdaBoost | 3.15% | 6.07% | 90.05% | 98.74% | 98.55% |

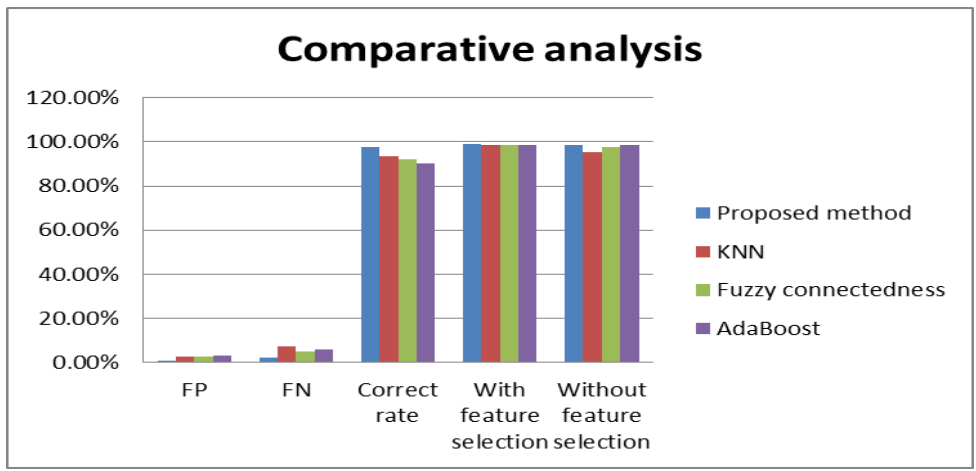

Fig: 5 Comparative analysis of existing algorithms and the proposed method

In this proposed system we used SVM for classification . Here we use two steps for classification one is SVM without continuous training another one is SVM with continuous training. The corresponding outputs are shown in figure 6 and figure 7

.Using this process we can easily identify the classification process and the accuracy. Continuous training gives more identification of the similar properties.

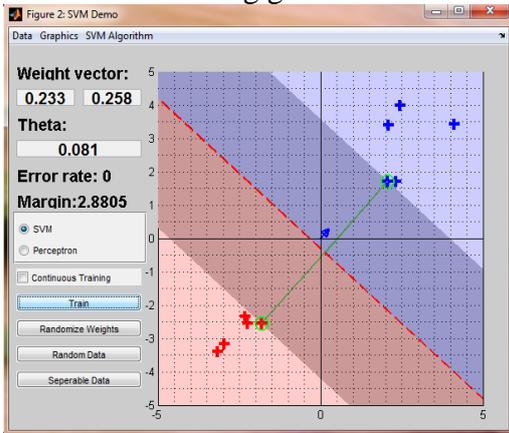
(a)

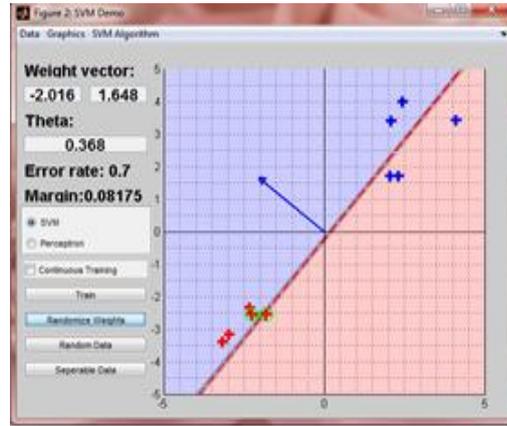
(b)

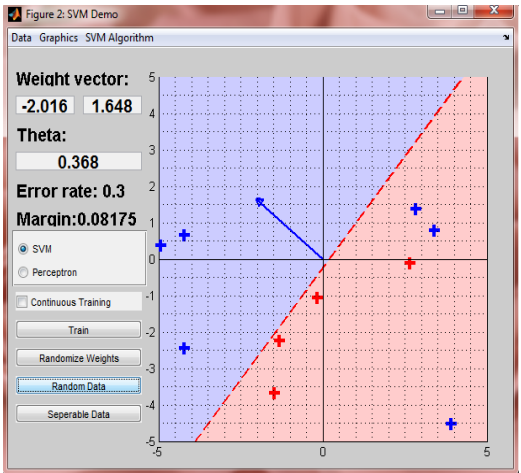
(c )

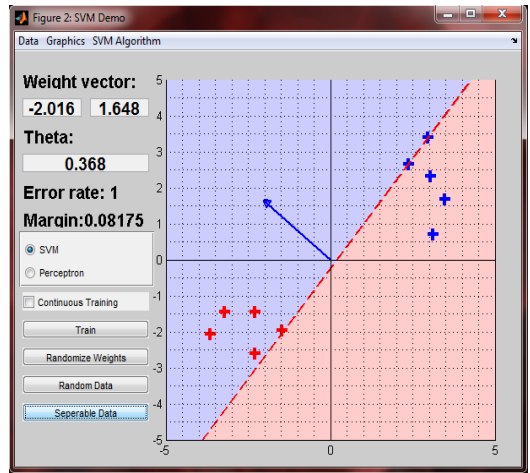
(d)

Fig 6 SVM without continuous training: Random weights, Random data and Sepeable data

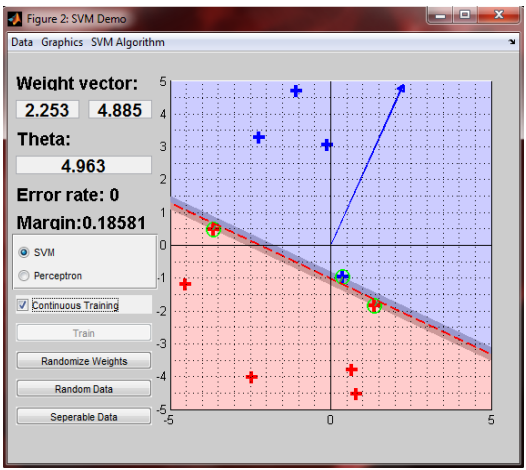
(a)

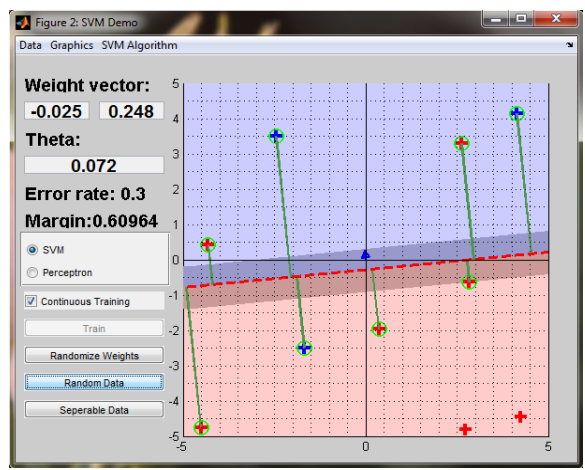
(b)

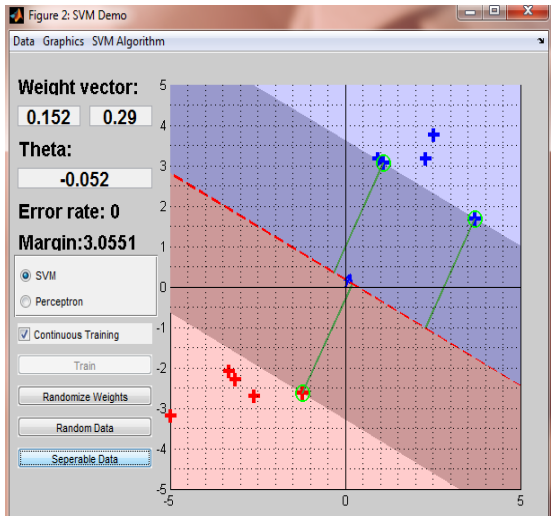 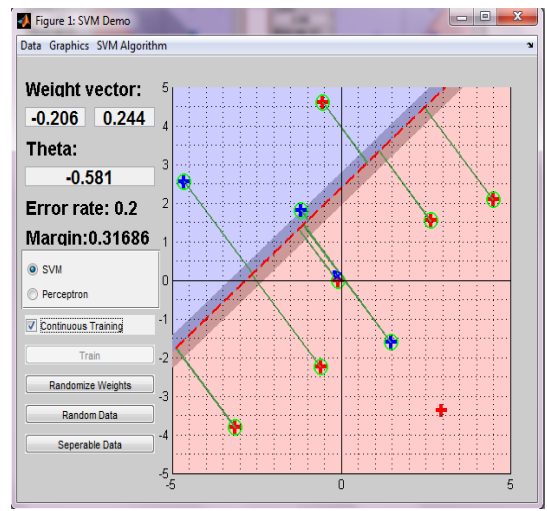

( C )                                                              ( d )

Fig 7. SVM with continuous training: Random weights, Random data and Seperable data

## 5. CONCLUSIONS

Brain Tumor MRI image Classification with feature selection and extraction have been carried out in the past with limited successs. The method suggested in this paper for the above work includes the steps, Image collection, Normalization, Intensity, shape and Texture feature extraction, feature selection and classification. In this method the shape, Intensity and Texture features are extracted and used for classification. Vital features are selected using LDA. The results are compared with PCA dimension reduction techniques. The number of features selected or features extracted by PCA and the classification accuracy by SVM is 98.87%. In this method we train the system by both continuous and without continuous data. So we minimize the error rate as well as increase the classification accuracy. Thus the proposed method performs better than the existing works. It is expected that the information of new imaging technique fMRI and the Image MOMENTS when added into the scheme will give more accurate results.

**ACKNOWLEDGEMENTS**

The work done by V.P.Gladis Pushpa Rathi, Dr. S.Palani is supported by Sudharsan Engineering College Sathiyamangalam, Pudukkottai, India

V.P.GladisPushpaRathi graduated from Cape Institute of Technology, Tirunelveli, did her post graduate studies at M.S.University Tirunelveli and is now doing her Ph.D in AnnaUniversity,Trichy. She is a faculty member of the department of Computer Science and Engineering, Sudharsan Engineering College. She has 5 years of teaching experience. Her field of Interest includes Digital Image Processing, Soft computing, and Datamining.

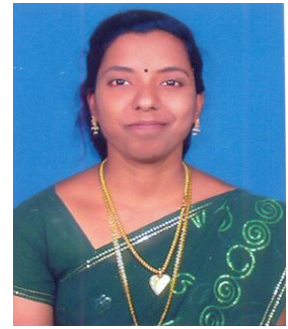

Dr.S Palani graduated from P.S.G. College of Technology, Coimbatore, did his post graduate studies at IIT Kharagpur and the doctoral degree from Regional Engineering College Trichy. He is a faculty member of the department of Electrical and Electronics Engineering, Sudharsan Engineering College, Pudukkottai, India. He has more than 40 years of teaching experience. His field of Interest includes Control Systems, Electrical Engineering and Digital signal Processing.

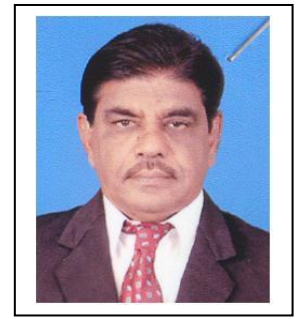

.